\documentclass[10pt,journal,letterpaper]{IEEEtran}
\usepackage{amsmath,amssymb,amsthm}
\usepackage{graphicx,cite,xcolor}
\usepackage[caption=false,font=footnotesize]{subfig}
\usepackage[hidelinks]{hyperref}
\usepackage{ulem}
\usepackage{nicefrac}
\ifdefined\pdfobjcompresslevel\pdfobjcompresslevel=0\fi

\newcommand{\R}{\mathbb{R}}

\newcommand{\T}{^{\mathsf T}}
\newcommand{\norm}[1]{\lVert #1\rVert}
\newif\ifdraftnotes
\draftnotestrue

\newtheorem{proposition}{Proposition}
\newtheorem{theorem}{Theorem}

\theoremstyle{remark}
\newtheorem{remark}{Remark}
\theoremstyle{plain}

\graphicspath{{figures/}}
\hypersetup{pdftitle={Directional Kinetic Energy Limitation for Physical Human-Robot Interaction}}

\AtBeginDocument{\normalem}

\begin{document}

\title{Safe and Effective Physical Human-Robot Interaction via \\ Directional Kinetic Energy Limitation}
\author{Riccardo Zanella, Wesley Roozing and Federico Califano%
\thanks{The authors are with Robotics and Mechatronics (RaM),
University of Twente, Enschede, The Netherlands.
Corresponding author: Riccardo Zanella
(e-mail: \href{mailto:schol@r-zanella.eu}{schol@r-zanella.eu}).}}
\maketitle

\begin{abstract}   
In physical human-robot interaction, limiting the robot's kinetic energy is a common strategy to ensure contact safety. For an anticipated contact direction, the kinetic energy associated with motion toward the contact may account for only a fraction of the robot's total kinetic energy. A bound on total kinetic energy also restricts complementary motion and retreat, potentially slowing task execution. In this letter, we introduce a one-sided directional kinetic-energy limit that isolates the approach component through a dynamically consistent decomposition of the robot's kinetic energy. We enforce this limit using a control barrier function in a quadratic program that computes the smallest admissible correction to the nominal joint torques. We show that, for a given energy budget, the directional bound can admit a wider range of robot velocities while retaining the prescribed limit on approach energy. Four experiments on a redundant manipulator demonstrate selective regulation of motion toward the anticipated contact. At the same energy budget, the proposed filter achieves approximately four times the peak end-effector speed of a total kinetic-energy filter.
\end{abstract}


\section{Introduction}
\label{sec:introduction}

Ensuring human safety is a primary concern in physical human-robot interaction (pHRI). However, achieving this goal is challenging because safety strongly depends on the interaction context: including the robot configuration, contact point and direction, human body region, contact geometry, and whether the body can move away \cite{zanella2026physical}.
The power and force limiting (PFL) framework of ISO/TS 15066 \cite{iso2016collaborative}, also addressed in ISO 10218-2 \cite{iso2025robotics}, provides body-region-specific force and pressure limits for assessing human-robot contacts. These limits are mapped to robot-motion constraints through a scalar mass-spring model, in which lumped robot and human masses interact through a compliant element along the contact normal. In this model, relative motion toward the contact drives tissue compression: the associated pre-impact kinetic energy determines the predicted peak force \cite{iso2016collaborative,zanella2026physical}. Thus, the relevant quantity is the kinetic energy associated with approach along the contact direction.

Energy-based safety has been investigated in literature mainly through reference-trajectory regulation \cite{laffranchi2009safe} and impedance adaptation \cite{tadele2014combining,raiola2018development}. A pre-collision strategy combines trajectory generation with constrained optimization to limit energy dissipated in potential impacts, while exploiting redundancy to reduce reflected mass \cite{rossi2015precollision}. Other formulations introduce transferred-energy indicators as quadratic constraints, adapting trajectory execution according to energy limits and human-robot separation \cite{meguenani2015control}, and extend this approach to account for post-impact loading \cite{meguenani2016energy}. While establishing the value of contact-aware energy regulation, these approaches cannot be transferred to an arbitrary artificial intelligence (AI) controller that exposes only its control actions.
However, the rapid development of and growing investment in AI-driven robot control demand safety mechanisms that can supervise different policies without accessing their internal structure or requiring retraining for each safety constraint.

\begin{figure}[t]
 \centering
 \includegraphics[width=0.85\columnwidth,trim=5mm 0mm 0mm 5mm,clip]{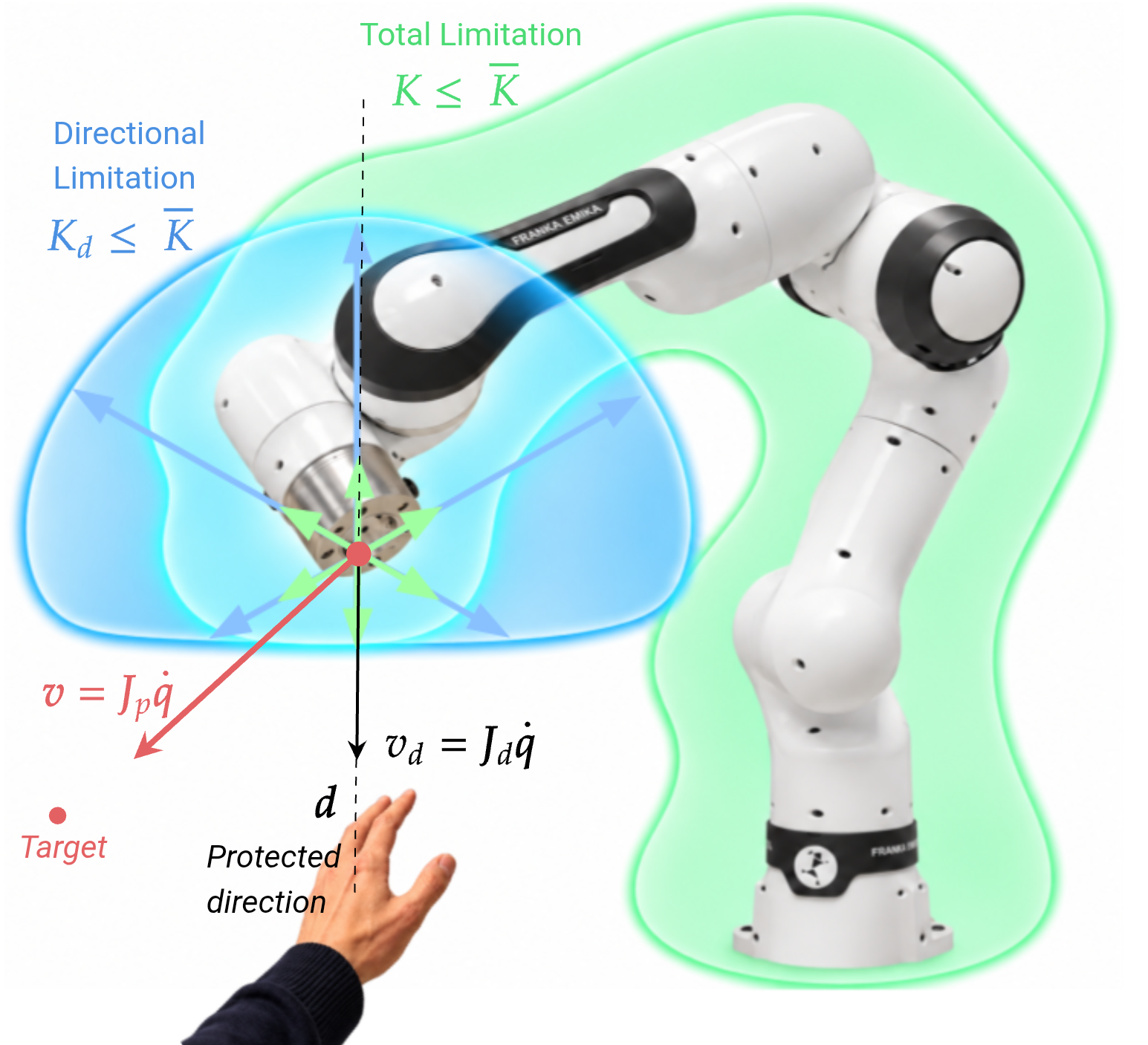}
 \caption{Conceptual comparison of total (green) and directional (blue)
 kinetic energy limits. The red arrow shows the end-effector velocity
 $\boldsymbol{v}$ toward the target. The protected unit direction
 $\boldsymbol{d}$ points toward the hand. The translational Jacobian
 $J_p$ maps joint velocities $\dot{\boldsymbol{q}}$ to $\boldsymbol{v}$, while
 $J_d=\boldsymbol{d}\T J_p$ maps them to the velocity component
 $v_d=\boldsymbol{d}\T\boldsymbol{v}$ along $\boldsymbol{d}$.
 The total formulation applies the upper bound $\bar K$ to the
 robot's total kinetic energy $K$. The directional formulation
 applies the same bound to $K_{d_+}$, which uses only the positive
 part of $v_d$ and therefore counts motion toward the hand.}
 \label{fig:concept} 
\end{figure}

Control barrier functions (CBFs) provide such a safety-filter interface: they encode admissible states through a barrier function and constrain the applied input to preserve those states \cite{ames2019control}. A quadratic program (QP) returns an admissible nominal command unchanged and otherwise minimizes its correction in a chosen input metric. This construction uses the robot model and the safety specification without interpreting or differentiating the nominal policy.
Most of the robotics literature applies CBFs to workspace or kinematic constraints, such as obstacle avoidance \cite{rauscher2016constrained,singletary2022safety}, whereas a smaller body of work addresses the energetic mechanisms of physical contact. Relevant formulations combine safety with passivity \cite{cortez2021safe}, characterize the energy transfer induced by barrier intervention \cite{califano2024effect}, or use CBFs to regulate energy tanks and power exchange \cite{michel2024novel}. The closest work, \cite{califano2025limiting}, directly limits the total kinetic energy of a torque-controlled robot through a CBF. However, this bound accounts for all kinetic-energy components, irrespective of their contribution to a particular contact.
Although the QP minimizes the instantaneous input correction, the resulting restriction on task performance also depends on the states excluded by the barrier.

A total kinetic-energy bound can therefore remain conservative even when enforced through a minimally modifying safety filter.
For example, during a handover, the human hand may lie along the robot's approach direction. Limiting the kinetic energy associated with motion toward the hand addresses the anticipated contact, whereas lateral corrections, retreat, and redundant joint motion need not be subject to the same bound.
Including these contributions in a single total kinetic-energy budget can unnecessarily slow the handover or restrict secondary tasks.
This motivates the present work, which limits the kinetic energy associated with approach toward a specified contact while leaving complementary motion and retreat outside the same budget, as illustrated in \autoref{fig:concept}.

When addressing safety in human-robot interaction through energy limitation, several approaches in the literature rely on virtual energy tanks \cite{califano2022use}. These tools can supervise a nominal controller through its power exchange with the robot, without requiring access to its internal structure, making them suitable for learned policies as well \cite{zanella2024learning}. However, many formulations are embedded within variable impedance or hierarchical control architectures, relying on specific controller components and their energetic behavior \cite{ferraguti2013tank,dietrich2017passive,michel2022safety}. This dependence limits their direct applicability to arbitrary AI-driven policies that expose only their control actions. Moreover, many applications primarily seek to preserve passivity \cite{ferraguti2013tank,dietrich2017passive}, which must not be equated with contact safety \cite{califano2022use}: passivity constrains net energy generation relative to the initial storage and external supply, but does not itself impose a contact-specific limit on the energy potentially transferred to a human. A smaller body of work develops safety-oriented tank formulations that limit energy injection according to kinetic-energy budgets \cite{michel2022safety}, enforce bounds motivated by ISO/TS 15066 \cite{benzi2023energy}, or combine tanks with CBFs to regulate energy and power \cite{michel2024novel}. Nevertheless, these formulations do not explicitly isolate the directional kinetic energy associated with approach toward a specified contact. Applying their budgets to a broader kinetic-energy quantity can therefore restrict motion unrelated to that contact, introducing conservatism similar to that of CBF-based total kinetic-energy limitation.

The proposed solution acts as a safety filter, independently of the nominal controller's internal structure, and exploits the anticipated contact direction to avoid unnecessary restrictions on robot motion. 

The main contributions of this letter are:
\begin{enumerate}
 \item A directional kinetic-energy limitation based on a dynamically consistent task and null-space decomposition, with separate  treatment of approach, complementary motion, and retreat. 
 \item CBF-based enforcement of the directional kinetic-energy constraint, thereby permitting higher velocities within the admissible bounds.
 \item Four experiments assessing reduced conservatism relative to total kinetic-energy limitation, preservation of motion outside the protected approach, and reduction of transient contact-force peaks.
\end{enumerate}

The remainder of this paper is organized as follows. Section~\ref{sec:background} recalls the CBF framework and total kinetic-energy limitation. Section~\ref{sec:methods} introduces the directional formulation and its theoretical analysis. Section~\ref{sec:experiments} presents four experiments, followed by a discussion in Section~\ref{sec:discussion} and conclusions in Section~\ref{sec:conclusion}.
\section{Background}
\label{sec:background}

\subsection{Control Barrier Functions}
\label{sec:cbf_background}

Consider the input-affine system $\dot{\boldsymbol{\xi}}=\boldsymbol{f}(\boldsymbol{\xi})+\mathbf{G}(\boldsymbol{\xi})\boldsymbol{u}$, with $\boldsymbol{\xi}\in\mathcal D\subseteq\R^N$, where $\mathcal D$ is open, $\boldsymbol{f}$ and $\mathbf{G}$ are locally Lipschitz, and $\boldsymbol{u}\in\mathcal U\subseteq\R^m$, with $\mathcal U$ nonempty, closed, and convex. Let $h:\mathcal D\to\R$ be continuously differentiable and define $\mathcal S_h:=\{\boldsymbol{\xi}\in\mathcal D:h(\boldsymbol{\xi})\ge0\}$, with $\nabla h(\boldsymbol{\xi})\ne0$ whenever $h(\boldsymbol{\xi})=0$.

The function $h$ is a control barrier function (CBF) if there exists a locally Lipschitz extended class-$\mathcal K$ function $\alpha$\footnote{A function $\alpha:(-b,a)\to\R$, with $a,b>0$, is of extended class $\mathcal K$ if it is continuous, strictly increasing, and satisfies $\alpha(0)=0$.} such that, for every $\boldsymbol{\xi}\in\mathcal D$, an input $\boldsymbol{u}\in\mathcal U$ satisfies
\begin{equation}
 \psi(\boldsymbol{\xi},\boldsymbol{u}):=
 \underbrace{\nabla h(\boldsymbol{\xi})\T
 \bigl(\boldsymbol{f}(\boldsymbol{\xi})+\mathbf{G}(\boldsymbol{\xi})\boldsymbol{u}\bigr)}_{\dot h(\boldsymbol{\xi},\boldsymbol{u})}
 +\alpha\bigl(h(\boldsymbol{\xi})\bigr)\ge0.
 \label{eq:cbf_constraint}
\end{equation}
Throughout this letter, we use the linear choice $\alpha(h):=\gamma h$, with $\gamma>0$.

\begin{theorem}[Forward invariance \cite{ames2019control}]
\label{thm:cbf_invariance}
Let $h$ be a CBF on $\mathcal D$. Any locally Lipschitz feedback $\boldsymbol{u}=\boldsymbol{k}(\boldsymbol{\xi})\in\mathcal U$ satisfying $\psi(\boldsymbol{\xi},\boldsymbol{k}(\boldsymbol{\xi}))\ge0$ for every $\boldsymbol{\xi}\in\mathcal D$ renders $\mathcal S_h$ forward invariant. Specifically, $\boldsymbol{\xi}(0)\in\mathcal S_h$ implies $\boldsymbol{\xi}(t)\in\mathcal S_h$ for all times for which the closed-loop solution exists in $\mathcal D$.
\end{theorem}

Since \eqref{eq:cbf_constraint} is affine in $\boldsymbol{u}$, it can be enforced through the quadratic program (QP)
\begin{equation}
 \begin{aligned}
 \boldsymbol{u}^\star(\boldsymbol{\xi}):=\arg\min_{\boldsymbol{u}\in\mathcal U}\;&
 \norm{\boldsymbol{u}-\boldsymbol{u}_{\rm nom}(\boldsymbol{\xi})}^2\\
 \text{subject to }& \psi(\boldsymbol{\xi},\boldsymbol{u})\ge0,
 \end{aligned}
 \label{eq:qp}
\end{equation}
where $\boldsymbol{u}_{\rm nom}$ is the nominal command. Whenever \eqref{eq:qp} is feasible, its optimizer is unique. If the resulting feedback is locally Lipschitz on $\mathcal D$, Theorem~\ref{thm:cbf_invariance} applies. Any nominal command satisfying both the input and CBF constraints is returned unchanged; otherwise, the filter selects the nearest feasible input in the Euclidean norm.

The result extends to continuously differentiable time-dependent
barriers $h(t,\boldsymbol{\xi})$ by using the total derivative
\begin{equation*}
 \dot h(t,\boldsymbol{\xi},\boldsymbol{u}):=
 \frac{\partial h}{\partial t}
 +\nabla_\xi h(t,\boldsymbol{\xi})\T\bigl(\boldsymbol{f}(\boldsymbol{\xi})+\mathbf{G}(\boldsymbol{\xi})\boldsymbol{u}\bigr).
\end{equation*}
Under the corresponding feasibility and regularity assumptions, a locally Lipschitz feedback $\boldsymbol{u}=\boldsymbol{k}(t,\boldsymbol{\xi})$ enforcing $\dot h+\alpha(h)\ge0$ preserves membership in $\mathcal S_h(t):=\{\boldsymbol{\xi}\in\mathcal D:h(t,\boldsymbol{\xi})\ge0\}$ for initially admissible states.

\subsection{Robot Dynamics and Total Kinetic Energy}

Consider a fully actuated, torque-controlled robot with
$n$ degrees of freedom (DoF), configuration $\boldsymbol{q}\in\mathcal Q$, and joint velocity $\dot{\boldsymbol{q}}\in\R^n$, governed by
\begin{equation}
 \mathbf{M}(\boldsymbol{q})\ddot{\boldsymbol{q}}+\mathbf{C}(\boldsymbol{q},\dot{\boldsymbol{q}})\dot{\boldsymbol{q}}+\boldsymbol{g}(\boldsymbol{q})
 =\boldsymbol{\tau}_{\rm act}+\boldsymbol{\tau}_{\rm ext},
 \label{eq:dynamics}
\end{equation}
where $\mathbf{M}=\mathbf{M}\T\succ0$ is the inertia matrix, $\mathbf{C}\dot{\boldsymbol{q}}$ is the Coriolis and centrifugal torque, $\boldsymbol{g}$ is the gravity torque, and $\boldsymbol{\tau}_{\rm act}$ and $\boldsymbol{\tau}_{\rm ext}$ are the actuator and external joint torques, respectively. 
We choose $\mathbf{C}$ such that $\dot{\mathbf{M}}-2\mathbf{C}$ is skew-symmetric.

The total kinetic energy is
\begin{equation}
 K(\boldsymbol{q},\dot{\boldsymbol{q}}):=\frac12\dot{\boldsymbol{q}}\T \mathbf{M}(\boldsymbol{q})\dot{\boldsymbol{q}}.
 \label{eq:total_kinetic_energy}
\end{equation}

\subsection{Limiting Total Kinetic Energy via CBFs}

As shown in \cite{califano2025limiting}, an upper bound $\bar K>0$ on the robot's total kinetic energy \eqref{eq:total_kinetic_energy} can be enforced through the barrier
\begin{equation*}
 h_{\text{tot}}(\boldsymbol{q},\dot{\boldsymbol{q}}):=\bar K-K(\boldsymbol{q},\dot{\boldsymbol{q}}),
 \qquad
 \mathcal S_{\text{tot}}
 :=\{(\boldsymbol{q},\dot{\boldsymbol{q}}):h_{\text{tot}}\ge0\}.
\end{equation*}
Using \eqref{eq:dynamics} and the skew-symmetry of $\dot{\mathbf{M}}-2\mathbf{C}$ gives $\dot h_{\text{tot}}
=-\dot{\boldsymbol{q}}\T(\boldsymbol{\tau}_{\rm act}+\boldsymbol{\tau}_{\rm ext}-\boldsymbol{g}(\boldsymbol{q}))$.
Hence, the corresponding CBF condition is
\begin{equation}
 \begin{aligned}
 \psi_{\text{tot}}(\boldsymbol{q},\dot{\boldsymbol{q}},\boldsymbol{\tau}_{\rm act})
 &:=-\dot{\boldsymbol{q}}\T(\boldsymbol{\tau}_{\rm act}+\boldsymbol{\tau}_{\rm ext}-\boldsymbol{g}(\boldsymbol{q}))\\
 &\quad+\gamma_{\text{tot}}h_{\text{tot}}\ge0,
 \end{aligned}
 \label{eq:cbf_constraint_tot}
\end{equation}
with $\gamma_{\text{tot}}>0$.
The total kinetic-energy bound can be enforced through \eqref{eq:qp} by taking $\boldsymbol{\tau}_{\rm act}$ as the input and substituting $\psi_{\text{tot}}$ for $\psi$.

On a regular model domain, $h_{\text{tot}}$ is continuously differentiable. Moreover, $h_{\text{tot}}=0$ implies $K=\bar K>0$, hence $\dot{\boldsymbol{q}}\ne0$ and $\partial h_{\text{tot}}/\partial\dot{\boldsymbol{q}}=-\dot{\boldsymbol{q}}\T \mathbf{M}\ne0$. Therefore, on any such domain where the QP remains feasible and its optimizer defines a locally Lipschitz feedback satisfying \eqref{eq:cbf_constraint_tot} for the actual dynamics, Theorem~\ref{thm:cbf_invariance} guarantees forward invariance of $\mathcal S_{\text{tot}}$.
Consequently, initially admissible states satisfy $K(t)\le\bar K$ while the closed-loop solution remains in that domain.

\section{Methods}
\label{sec:methods}

In this section we construct a safety filter that limits the kinetic energy associated with approach along a specified contact direction. We first use a dynamically consistent task and null-space decomposition to isolate this contribution, then distinguish approach from retreat and derive the corresponding torque-affine CBF constraint. Finally, we compare its admissible states and velocities with those of the total-energy limitation to quantify the reduction in conservatism.

\subsection{Kinetic Energy Decomposition}  
Let $\boldsymbol{\nu}:=\mathbf{J}(\boldsymbol{q})\dot{\boldsymbol{q}}$ describe a geometric velocity task of dimension $\ell$.
For the end-effector pose task, its associated velocity $\boldsymbol{\nu}:=[\boldsymbol{v}\T\ \boldsymbol{\omega}\T]\T$ contains linear and angular velocity in a common frame. Following \cite{califano2025geometric}, on a domain where $\operatorname{rank}\mathbf{J}=\ell$, we define
\begin{equation}
 \boldsymbol{\Lambda}:=(\mathbf{J}\mathbf{M}^{-1}\mathbf{J}\T)^{-1},\qquad \mathbf{J}_M^\dagger:=\mathbf{M}^{-1}\mathbf{J}\T\boldsymbol{\Lambda}.
 \label{eq:operational}
\end{equation}
Here $\boldsymbol{\Lambda}$ is the task-space inertia matrix, and $\mathbf{J}_M^\dagger$ is the dynamically consistent inverse of $\mathbf{J}$, satisfying $\mathbf{J}\mathbf{J}_M^\dagger=\mathbf{I}_\ell$.
It maps a task velocity to its minimum-kinetic-energy joint realization.

The dynamically consistent task and null-space decomposition is $\dot{\boldsymbol{q}}=\dot{\boldsymbol{q}}_{\rm task}+\dot{\boldsymbol{q}}_{\rm null}$, with
\begin{equation}
 \dot{\boldsymbol{q}}_{\rm task}:=\mathbf{J}_M^\dagger \mathbf{J}\dot{\boldsymbol{q}},\qquad
 \dot{\boldsymbol{q}}_{\rm null}:=(\mathbf{I}-\mathbf{J}_M^\dagger \mathbf{J})\dot{\boldsymbol{q}}.
 \label{eq:vel_decomposition}
\end{equation}
The first component generates $\boldsymbol{\nu}$; the second satisfies $\mathbf{J}\dot{\boldsymbol{q}}_{\rm null}=0$ and does not contribute to the selected task velocity.
Since $\dot{\boldsymbol{q}}_{\rm task}\T \mathbf{M}\dot{\boldsymbol{q}}_{\rm null}=0$, kinetic energy decomposes as
\begin{align*}
 K&=\tfrac12\dot{\boldsymbol{q}}\T \mathbf{M}\dot{\boldsymbol{q}}=K_{\rm task}+K_{\rm null},\\
 K_{\rm task}&:=\tfrac12\boldsymbol{\nu}\T\boldsymbol{\Lambda}\boldsymbol{\nu},\qquad
 K_{\rm null}:=\tfrac12\dot{\boldsymbol{q}}_{\rm null}\T \mathbf{M}\dot{\boldsymbol{q}}_{\rm null}.
\end{align*}
Thus, omitting null-space velocity in $\dot{\boldsymbol{q}}=\mathbf{J}_M^\dagger\boldsymbol{\nu}$ would exclude kinetic energy that can be present in a redundant robot \cite{califano2025geometric}.  

\subsection{Directional Kinetic Energy}
\label{sec:directional_energy}


Let $\mathbf{J}_p$ denote the end-effector translational Jacobian, so that $\boldsymbol{v}:=\mathbf{J}_p\dot{\boldsymbol{q}}$ is the linear velocity of the end-effector origin in the base frame. For a unit direction $\boldsymbol{d}$, with $\norm{\boldsymbol{d}}=1$, we define
\begin{equation}
 \mathbf{J}_d:=\boldsymbol{d}\T \mathbf{J}_p,\quad v_d:=\mathbf{J}_d\dot{\boldsymbol{q}},\quad
 m_d:=(\mathbf{J}_d\mathbf{M}^{-1}\mathbf{J}_d\T)^{-1}.
 \label{eq:directional_task}
\end{equation}
We restrict the construction to a regular domain where $\mathbf{J}_d\ne0$, so that $m_d$ is finite and positive.
As illustrated in \autoref{fig:concept}, the directional Jacobian $\mathbf{J}_d$ selects the scalar projection $v_d=\boldsymbol{d}\T\boldsymbol{v}$ of the end-effector velocity along $\boldsymbol{d}$. It defines a scalar task, corresponding to $\ell=1$ and $\mathbf{J}=\mathbf{J}_d$ in \eqref{eq:operational}. Hence, the task-space inertia $\boldsymbol{\Lambda}$ reduces to the reflected inertia $m_d$, and the dynamically consistent inverse is $\mathbf{J}_{d,M}^\dagger:=\mathbf{M}^{-1}\mathbf{J}_d\T m_d$.

First we define the bilateral directional kinetic energy, which treats both senses along the selected axis equally:
\begin{equation*}
 K_d:=\tfrac12m_dv_d^2.
\end{equation*}
\begin{remark} 
Specializing \eqref{eq:vel_decomposition} to the scalar task $\mathbf{J}_d$ gives the task component $\dot{\boldsymbol{q}}_{\parallel,d}$ and complementary null-space component $\dot{\boldsymbol{q}}_{\perp,d}$ of the joint velocity:
\begin{align}
 \dot{\boldsymbol{q}}_{\parallel,d}&:=\mathbf{M}^{-1}\mathbf{J}_d\T m_dv_d,\qquad
 \dot{\boldsymbol{q}}_{\perp,d}:=\dot{\boldsymbol{q}}-\dot{\boldsymbol{q}}_{\parallel,d},\notag
\\
 K_d&=\tfrac12\dot{\boldsymbol{q}}_{\parallel,d}\T \mathbf{M}\dot{\boldsymbol{q}}_{\parallel,d},
 \qquad K=K_d+\tfrac12\dot{\boldsymbol{q}}_{\perp,d}\T \mathbf{M}\dot{\boldsymbol{q}}_{\perp,d}.
 \label{eq:scalar_decomposition}
\end{align}
Here, $\dot{\boldsymbol{q}}_{\parallel,d}$ produces $v_d$, whereas $\mathbf{J}_d\dot{\boldsymbol{q}}_{\perp,d}=0$.
These components are orthogonal with respect to $\mathbf{M}$; 
the subscripts do not denote Cartesian directions. At fixed $\boldsymbol{q}$, $\dot{\boldsymbol{q}}_{\parallel,d}$ minimizes kinetic energy among all candidate joint velocities $\boldsymbol{w}\in\R^n$ satisfying $\mathbf{J}_d\boldsymbol{w}=v_d$. Hence, complementary motion contributes to $K$ but not to $K_d$.

This identity supports the subsequent comparison with total limitation; the components need not be computed for the CBF. Note that exclusion from the budget does not guarantee that the torque correction leaves complementary motion unchanged.
\end{remark}

Since we aim to limit only the kinetic energy associated with motion toward a potential contact, we orient $\boldsymbol{d}$ toward that contact. Thus, $v_d>0$ corresponds to end-effector motion toward the contact, whereas $v_d<0$ corresponds to motion away from it. We therefore define
\begin{equation}
 v_{d_+} := \max(v_d,0),
 \qquad
 K_{d_+} := \frac12 m_d v_{d_+}^2.
 \label{eq:unilateral}
\end{equation}
Then, $K_{d_+}=K_d$ during approach, whereas $K_{d_+}=0$ during retreat. Note that this definition leaves retreat unconstrained by the directional kinetic-energy bound. Moreover, the square of the positive-part function is continuously differentiable, so $K_{d_+}$ is continuously differentiable on a regular domain, including at $v_d=0$.

\subsection{Limiting Directional Kinetic Energy via CBFs}
For a potential contact with a human body region along $\boldsymbol{d}$, the directional kinetic energy $K_{d_+}$ captures the contribution associated with motion toward the contact ($v_d>0$), as illustrated in \autoref{fig:concept}. To bound this contribution, we introduce a CBF with a constant upper bound $\bar K>0$.
Ideally, this bound should be derived from the ISO force and pressure limits for the relevant body region under the assumed contact conditions \cite{iso2016collaborative,zanella2026physical}.
Specifically, we define
\begin{equation*}
 h_d(\boldsymbol{q},\dot{\boldsymbol{q}}):=\bar K-K_{d_+}(\boldsymbol{q},\dot{\boldsymbol{q}}),\qquad
 \mathcal S_d:=\{(\boldsymbol{q},\dot{\boldsymbol{q}}):h_d\ge0\}.
\end{equation*}
Differentiating \eqref{eq:unilateral}, we obtain
\begin{equation}
 \dot K_{d_+}=m_dv_{d_+}\dot v_d+\tfrac12\dot m_dv_{d_+}^2,
 \label{eq:energy_derivative}
\end{equation}
where the directional acceleration and reflected-inertia derivative are
\begin{align*}
 \dot v_d&=\dot{\mathbf{J}}_d\dot{\boldsymbol{q}}+\mathbf{J}_d\ddot{\boldsymbol{q}},
\\
 \dot m_d&=-m_d^2\bigl(2\dot{\mathbf{J}}_d\mathbf{M}^{-1}\mathbf{J}_d\T
 -\mathbf{J}_d\mathbf{M}^{-1}\dot{\mathbf{M}} \mathbf{M}^{-1}\mathbf{J}_d\T\bigr).
\end{align*}
The latter follows by differentiating the reciprocal directional mobility in \eqref{eq:directional_task}. The inertia derivative can be obtained from the model as $\dot{\mathbf{M}}=\sum_i(\partial \mathbf{M}/\partial q_i)\dot q_i$, or from $\dot{\mathbf{M}}=\mathbf{C}+\mathbf{C}\T$ for a matrix $\mathbf{C}$ satisfying the stated identity.
For a fixed $\boldsymbol{d}$, we have $\dot{\mathbf{J}}_d=\boldsymbol{d}\T\dot{\mathbf{J}}_p$. For a prescribed, smoothly time-varying direction $\boldsymbol{d}(t)$, we instead use $\dot{\mathbf{J}}_d=\dot{\boldsymbol{d}}\T \mathbf{J}_p+\boldsymbol{d}\T\dot{\mathbf{J}}_p$. 

For physical actuator torque as input, combining \eqref{eq:dynamics} with \eqref{eq:energy_derivative} gives
\begin{align*}
 \dot{h}_d&=\boldsymbol{a}\boldsymbol{\tau}_{\rm act}+b_{\mathrm{phys}},\\
 \boldsymbol{a}&:=-v_{d_+}m_d\mathbf{J}_d\mathbf{M}^{-1},\\
 b_{\mathrm{phys}}&:=-v_{d_+}m_d\dot{\mathbf{J}}_d\dot{\boldsymbol{q}}-\tfrac12\dot m_dv_{d_+}^2\\
 &\quad+v_{d_+}m_d\mathbf{J}_d\mathbf{M}^{-1}(\mathbf{C}\dot{\boldsymbol{q}}+\boldsymbol{g}-\boldsymbol{\tau}_{\rm ext}).
\end{align*}

With gravity and Coriolis compensation, we define the residual input $\boldsymbol{u}:=\boldsymbol{\tau}_{\rm act}-\mathbf{C}\dot{\boldsymbol{q}}-\boldsymbol{g}$. Then $\mathbf{M}\ddot{\boldsymbol{q}}=\boldsymbol{u}+\boldsymbol{\tau}_{\rm ext}$ and
\begin{align*}
 \dot{h}_d&=\boldsymbol{a}\boldsymbol{u}+b,\qquad b:=b_0+b_{\mathrm{ext}},\\
 b_0&:=-v_{d_+}m_d\dot{\mathbf{J}}_d\dot{\boldsymbol{q}}-\tfrac12\dot m_dv_{d_+}^2,\\
 b_{\mathrm{ext}}&:=-v_{d_+}m_d\mathbf{J}_d\mathbf{M}^{-1}\boldsymbol{\tau}_{\rm ext}.
\end{align*}
In these input coordinates, the CBF constraint \eqref{eq:cbf_constraint} is
\begin{equation}
 \psi_d(\boldsymbol{q},\dot{\boldsymbol{q}},\boldsymbol{u}):=\boldsymbol{a}\boldsymbol{u}+b+\gamma_d h_d\ge0.
 \label{eq:cbf_constraint_dir}
\end{equation}
Here $\gamma_d>0$. To limit the directional kinetic energy $K_{d_+}$, we enforce \eqref{eq:cbf_constraint_dir} through \eqref{eq:qp} by substituting $\psi_d$ for $\psi$.
The nominal command and actuator limits are expressed in the same residual-input coordinates; the corresponding physical torque is $\boldsymbol{\tau}_{\rm act}=\boldsymbol{u}^\star+\mathbf{C}\dot{\boldsymbol{q}}+\boldsymbol{g}$.

Note that, when the end-effector moves away from the potential contact or has zero directional velocity ($v_d\le0$), we have $K_{d_+}=0$, $\boldsymbol{a}=0$, and $b=0$. Hence, \eqref{eq:cbf_constraint_dir} is automatically satisfied because $\gamma_d\bar K>0$.
Thus, this constraint does not require any modification of the nominal command during retreat. 

On a regular domain where $\mathbf{J}_d\ne0$, $h_d$ is continuously differentiable, including at $v_d=0$.
Moreover, $h_d=0$ implies $K_{d_+}=\bar K>0$, hence $v_d>0$ and $\partial h_d/\partial\dot{\boldsymbol{q}}=-m_dv_d\mathbf{J}_d\ne0$. 
Therefore, on any such domain where the QP remains feasible and its optimizer defines a locally Lipschitz feedback satisfying \eqref{eq:cbf_constraint_dir} for the actual dynamics, Theorem~\ref{thm:cbf_invariance} guarantees forward invariance of $\mathcal S_d$.
Consequently, initially admissible states satisfy $K_{d_+}(t)\le\bar K$ while the closed-loop solution remains in that domain. For a prescribed $\boldsymbol{d}(t)$, the barrier and its admissible set depend explicitly on time, and the corresponding extension in Section~\ref{sec:cbf_background} applies.

\subsection{Comparison with Total Kinetic-Energy Limitation}
\label{sec:comparison}
We compare the kinetic-energy bounds enforced by \eqref{eq:cbf_constraint_dir} and \eqref{eq:cbf_constraint_tot} in terms of their admissible states and velocities. This comparison quantifies the additional restrictions introduced by a total kinetic-energy bound when the selected approach component is the quantity of interest.

On a common regular domain, for the same budget $\bar K>0$, we have $0\le K_{d_+}\le K$, and hence
\begin{equation}
 \mathcal S_{\text{tot}}
 \subseteq \mathcal S_d.
 \label{eq:set_inclusion}
\end{equation}
Thus, every state satisfying the total bound also satisfies the directional bound. The converse does not hold in general: complementary motion or retreat can yield $K>\bar K$ while $K_{d_+}\le\bar K$. Therefore, the directional limitation admits states excluded by the total-energy limitation while retaining the selected directional kinetic-energy bound. Note that this comparison concerns admissible states; the instantaneous CBF constraints also depend on kinetic-energy derivatives.

To quantify the extra velocity margin provided by the directional kinetic-energy bound, we fix a regular configuration $\boldsymbol{q}$ and consider $\dot{\boldsymbol{q}}:=s\boldsymbol{w}$, with $\boldsymbol{w}$ fixed, $\mathbf{J}_d\boldsymbol{w}>0$, and $s\ge0$. Increasing $s$ increases the joint-speed magnitudes while preserving their relative components.

\begin{proposition}[Larger admissible velocity]
\label{prop:velocity}
For the same budget $\bar K>0$, let
\begin{equation*}
 s_{\text{tot}}^{\max}:=\sqrt{2\bar K/(\boldsymbol{w}\T \mathbf{M}\boldsymbol{w})},
 \qquad
 s_d^{\max}:=\sqrt{2\bar K/(m_d(\mathbf{J}_d\boldsymbol{w})^2)}
\end{equation*}
denote the maximum values of $s$ satisfying $K\le\bar K$ and $K_{d_+}\le\bar K$, respectively. Then
\begin{equation}
 s_d^{\max}\ge s_{\text{tot}}^{\max}.
 \label{eq:speed_comparison}
\end{equation} 
\end{proposition}

\begin{proof}
Define $\boldsymbol{w}_{\perp,d}:=(\mathbf{I}-\mathbf{J}_{d,M}^\dagger \mathbf{J}_d)\boldsymbol{w}$. Applying \eqref{eq:scalar_decomposition} to $\boldsymbol{w}$ gives $\boldsymbol{w}\T \mathbf{M}\boldsymbol{w}=m_d(\mathbf{J}_d\boldsymbol{w})^2+\boldsymbol{w}_{\perp,d}\T \mathbf{M}\boldsymbol{w}_{\perp,d}$. Since $\mathbf{M}\succ0$, the complementary term is nonnegative and vanishes if and only if $\boldsymbol{w}_{\perp,d}=0$. Comparing the denominators in the definitions of $s_{\text{tot}}^{\max}$ and $s_d^{\max}$ yields \eqref{eq:speed_comparison}.
\end{proof}

The end-effector speed satisfies $\norm{\mathbf{J}_p\dot{\boldsymbol{q}}}=s\norm{\mathbf{J}_p\boldsymbol{w}}$, with $\norm{\mathbf{J}_p\boldsymbol{w}}>0$ because $\mathbf{J}_d\boldsymbol{w}>0$. Consequently, the same ordering holds for the maximum end-effector speeds among the considered joint velocities.

\section{Experimental Results}
\label{sec:experiments}
To evaluate the practical performance of directional kinetic energy limiting, and to examine its potential advantages relative to limiting the total kinetic energy, we conducted four experiments. Where applicable, we compare nominal control without a safety filter, $\boldsymbol{u}_{\rm nom}$, the CBF limiting total kinetic energy through \eqref{eq:cbf_constraint_tot} \cite{califano2025limiting}, and the proposed CBF enforcing the directional bound in \eqref{eq:cbf_constraint_dir}. The first three experiments examine admissible motion and task performance; the fourth evaluates the resulting transient contact forces.

\subsection{Platform, Controller, and Protocol}

Experiments are performed on a seven-DoF Franka Emika Panda using a ROS-based joint-torque controller through the Franka Control Interface (FCI). The torque loop runs at 1~kHz, while the reference pose is updated at 30~Hz. The nominal command is generated by the Cartesian impedance law

\begin{equation*}
 \boldsymbol{u}_{\rm nom}:=-\mathbf{J}\T(\mathcal K \boldsymbol{e}+\mathcal B\boldsymbol{\nu}),
\end{equation*}
where $\boldsymbol{e}$ stacks the position and orientation errors expressed in the same frame as the geometric task velocity $\boldsymbol{\nu}:=\mathbf{J}\dot{\boldsymbol{q}}$, and $\mathcal K$ and $\mathcal B$ are the stiffness and damping matrices, respectively. We use this known and interpretable controller to facilitate comparisons across experimental conditions. The safety filter nevertheless operates on the nominal torque command, so the same interface can supervise an AI controller.

Gravity compensation is supplied by the platform, and Coriolis compensation is added after the CBF optimization. The Jacobian and inertia are evaluated from the robot model, with their time derivatives estimated from successive evaluations at the torque-loop period. All reported directions are expressed in the robot base frame.

\subsection{Experiment 1: Directional Versus Total Kinetic-Energy Limitation}

\begin{figure}[t]
 \centering
 \includegraphics[width=\columnwidth,trim=7mm 0mm 0mm 0mm,clip]{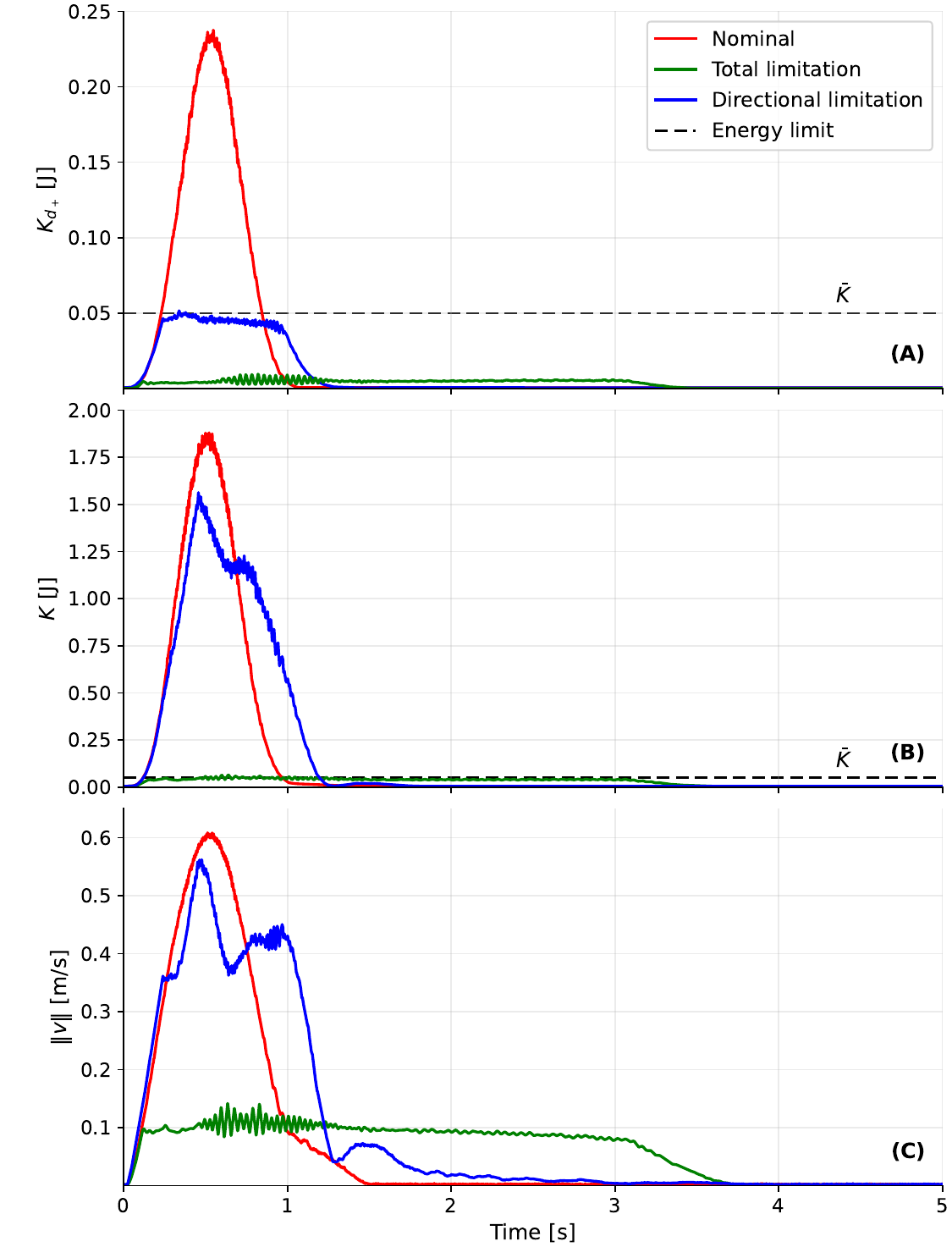}
 \caption{Target reaching with nominal control and CBF-based limitation
 of total or directional kinetic energy, with $+x$ as the protected
 direction.
 (A) Directional kinetic energy $K_{d_+}$ in the protected direction
 for all three control conditions. (B) Total kinetic energy $K$.
 (C) End-effector speed $\norm{\boldsymbol{v}}$.
 The dashed lines indicate $\bar K=0.05~\mathrm{J}$.
 The proposed filter permits substantially higher speed and
 total kinetic energy while regulating the protected approach.}

 \label{fig:target}
\end{figure}

In the first experiment, the robot performs target reaching predominantly along the positive $x$-axis. We assume a potential human contact along this approach and select $\boldsymbol{d}:=(1,0,0)\T$ as the protected direction. The proposed CBF bounds $K_{d_+}$, whereas the baseline CBF bounds $K$, both with the same budget $\bar K=0.05~\mathrm{J}$.

\autoref{fig:target}(A) and (B) show the directional kinetic energy $K_{d_+}$ and total kinetic energy $K$, respectively. With nominal control, their peaks are approximately $0.24~\mathrm{J}$ and $1.88~\mathrm{J}$. The baseline CBF keeps $K$ near the budget, while the peak of $K_{d_+}$ remains below $0.01~\mathrm{J}$, less than $20\%$ of the available directional budget. By contrast, the proposed CBF regulates $K_{d_+}$ near $0.05~\mathrm{J}$ while permitting $K$ to reach approximately $1.56~\mathrm{J}$. This behavior is consistent with \eqref{eq:scalar_decomposition} and \eqref{eq:set_inclusion}: complementary kinetic energy can exceed the directional budget without violating the selected directional bound.

The corresponding end-effector speed, $\norm{\boldsymbol{v}}=\norm{\mathbf{J}_p\dot{\boldsymbol{q}}}$, is shown in \autoref{fig:target}(C). Its peak is approximately $0.61~\mathrm{m/s}$ under nominal control, $0.56~\mathrm{m/s}$ with the proposed filter, and $0.14~\mathrm{m/s}$ with the total kinetic-energy limitation. Thus, directional limitation achieves a peak about four times that obtained by limiting total kinetic energy and retains approximately $92\%$ of the nominal peak. Moreover, the baseline filter spreads the motion over a longer interval, whereas the proposed filter preserves a faster reaching transient. These observations illustrate the velocity margin established in Proposition~\ref{prop:velocity}.

\subsection{Experiment 2: Changing the Constrained Direction}

We repeat the same target-reaching task, predominantly along the positive $x$-axis of the robot base frame, while selecting $\boldsymbol{d}_x:=(1,0,0)\T$, $\boldsymbol{d}_y:=(0,1,0)\T$, or $\boldsymbol{d}_{xy}:=(1,1,0)\T/\sqrt{2}$ as the protected direction for a potential human-robot contact. In all three cases, we use the same directional kinetic-energy budget $\bar K=0.05~\mathrm{J}$.

\begin{figure}[t]
 \centering
 \includegraphics[width=\columnwidth,trim=7mm 0mm 0mm 0mm,clip]{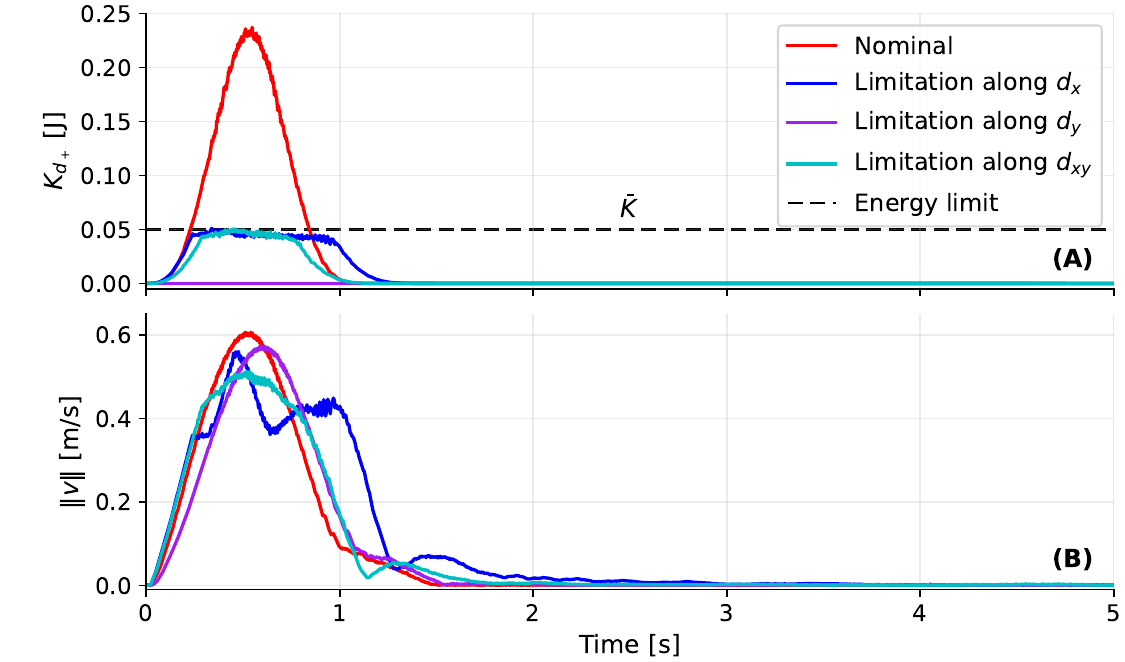}
 \caption{Target reaching with directional kinetic-energy limitation
 along $\boldsymbol{d}_x$, $\boldsymbol{d}_y$, or $\boldsymbol{d}_{xy}$ in separate trials.
 (A) Directional kinetic energy $K_{d_+}$ for each trial's protected
 direction; the nominal curve shows $K_{(d_x)_+}$.
 (B) End-effector speed $\norm{\boldsymbol{v}}$.
 The dashed line indicates $\bar K=0.05~\mathrm{J}$.
 The $\boldsymbol{d}_y$ constraint leaves the speed close to nominal because
 this predominantly $x$-directed task weakly excites its protected
 direction.}

 \label{fig:directions}
\end{figure}

\autoref{fig:directions}(A) shows that the protected directional kinetic energy $K_{d_+}$ reaches the budget for $\boldsymbol{d}_x$ and $\boldsymbol{d}_{xy}$, whereas it remains negligible for $\boldsymbol{d}_y$. This follows the scalar projection in \eqref{eq:directional_task}: motion along $+x$ contributes directly to $v_{d_x}$ and partially to $v_{d_{xy}}$, but contributes no velocity along the orthogonal direction $\boldsymbol{d}_y$. Consequently, the same task can require intervention for one potential contact direction and little or none for another. This is the selectivity absent from a total kinetic-energy bound, which counts all joint motion.
Consistently, \autoref{fig:directions}(B) shows that the $\boldsymbol{d}_y$ trial remains close to nominal, with a peak speed of approximately $0.57~\mathrm{m/s}$ compared with $0.61~\mathrm{m/s}$. Both the $\boldsymbol{d}_x$ and $\boldsymbol{d}_{xy}$ constraints reshape the reaching transient; the oblique-direction trial has a lower peak speed, approximately $0.51~\mathrm{m/s}$, but a shorter main transient than the $\boldsymbol{d}_x$ trial. Note that an oblique direction does not imply a response numerically intermediate between the other trials: reflected inertia and the nominal dynamics also determine the correction.

\subsection{Experiment 3: Circular Trajectory}

\begin{figure}[t]
 \centering
 \includegraphics[width=\columnwidth,trim=7mm 0mm 0mm 0mm,clip]{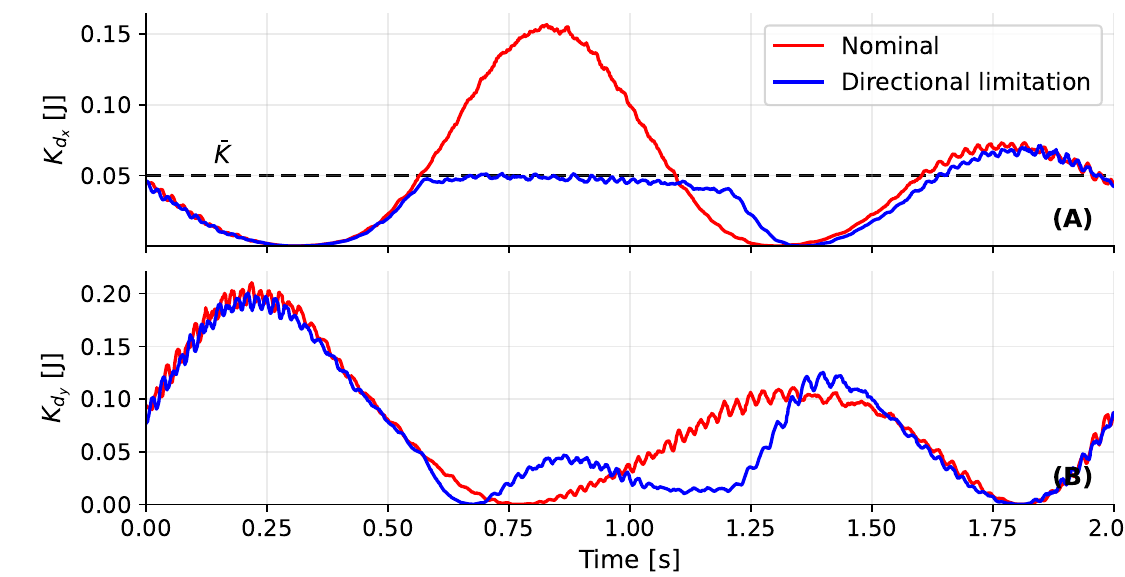}
 \caption{Circular trajectory with nominal control and the
 proposed CBF protecting $+x$. (A) Bilateral directional
 kinetic energy $K_{d_x}$ along the $x$-axis. (B) Bilateral directional
 kinetic energy $K_{d_y}$ along the $y$-axis. The dashed line marks
 $\bar K=0.05~\mathrm{J}$. Only the directional kinetic energy $K_{(d_x)_+}$
 is constrained: $K_{d_x}$ may exceed the budget during motion toward
 $-x$, and no bound is imposed on $K_{d_y}$.}

 \label{fig:circle}
\end{figure}

We next evaluate a circular reference trajectory in the horizontal plane, centered at $(x,y,z)=(0.43,0,0.50)~\mathrm{m}$, with a radius of $0.17~\mathrm{m}$ and a frequency of $0.5~\mathrm{Hz}$. \autoref{fig:circle} compares nominal control with the proposed CBF protecting $\boldsymbol{d}:=(1,0,0)\T$ over one 2~s revolution, using $\bar K=0.05~\mathrm{J}$.

The figure reports the bilateral directional energies $K_{d_x}$ and $K_{d_y}$, as defined in Section~\ref{sec:directional_energy}. These diagnostics include both senses of motion along each axis, whereas the CBF limits only $K_{(d_x)_+}$. During motion toward $+x$, \autoref{fig:circle}(A) shows the corresponding energy regulated near the budget instead of reaching the nominal peak of approximately $0.16~\mathrm{J}$. During retreat toward $-x$, the curves remain close and $K_{d_x}$ can exceed $0.05~\mathrm{J}$. This is consistent with \eqref{eq:unilateral}: retreat gives $K_{(d_x)_+}=0$ and does not require intervention by the proposed CBF.

The $y$-axis energy in \autoref{fig:circle}(B) likewise remains close to nominal over substantial portions of the revolution and is allowed to exceed the protected-direction budget. Its deviations occur mainly during and after the intervention along $+x$. They are consistent with the coupling between the torque correction and Cartesian motion: slowing the $x$ component changes the tracking error while the reference continues around the circle, which can also change the path and timing of the $y$ motion. Thus, leaving a component outside the constraint does not guarantee an identical closed-loop trajectory for that component. The figure demonstrates selective regulation of the directional kinetic energy $K_{(d_x)_+}$, with comparatively small changes away from the constrained phase.

\subsection{Experiment 4: Transient Contact}

\begin{figure}[t]
 \centering
 \includegraphics[width=0.85\columnwidth]{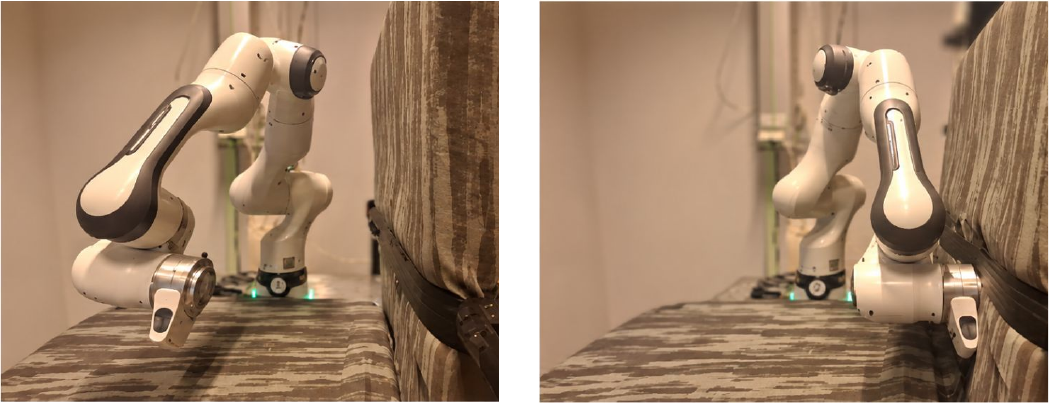}
 \caption{Transient-contact experiment setup. Left: approach toward the compliant object along the protected $+y$ direction. Right: contact.}
 \label{fig:setup}
\end{figure}

\begin{figure}[t]
 \centering
 \includegraphics[width=\columnwidth,trim=10mm 0mm 0mm 0mm,clip]{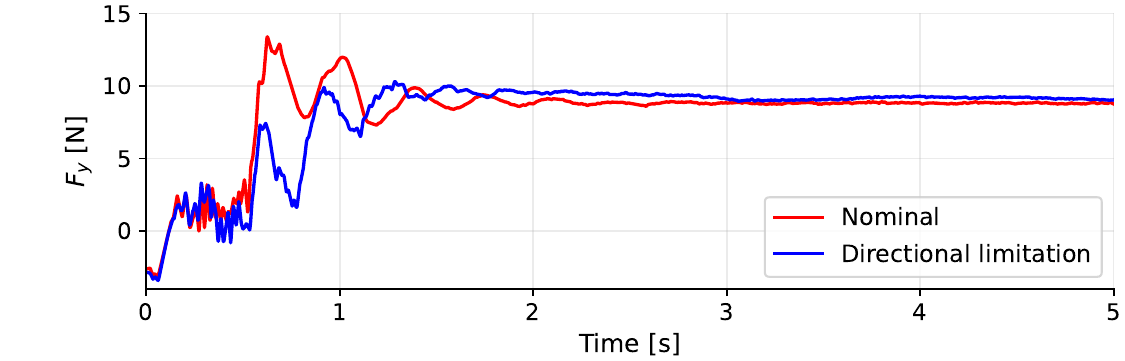}
 \caption{Estimated contact-direction force $F_y$ with nominal control and the proposed filter. The peak decreases from approximately $13.4~\mathrm{N}$ to $10.3~\mathrm{N}$, while sustained loading remains after the transient.}
 \label{fig:contact}
\end{figure}

In the final experiment, the robot approaches a compliant object along the positive $y$-axis, with the protected direction $\boldsymbol{d}:=(0,1,0)\T$ expressed in the robot base frame and a kinetic-energy budget $\bar K=0.25~\mathrm{J}$. \autoref{fig:setup} illustrates the contact task. We evaluate the interaction through the external Cartesian force estimate derived from the robot's internal joint-torque sensors and model. \autoref{fig:contact} reports its component $F_y$ along the contact direction for nominal control and the proposed filter in two separate trials.

The estimated force peak decreases from approximately $13.4~\mathrm{N}$ to $10.3~\mathrm{N}$, a reduction of about $23\%$. This observation connects directional kinetic-energy limitation to the force-based motivation of PFL: under the same effective contact masses and compliance, reducing the kinetic energy driving normal compression reduces the peak force predicted by the scalar contact model \cite{iso2016collaborative,zanella2026physical}. The experiment supports this mechanism through a lower estimated transient peak, although it does not establish a body-region-specific force or pressure guarantee. A sustained force of approximately $9~\mathrm{N}$ remains in both trials, illustrating that a pre-contact kinetic-energy bound does not regulate subsequent quasi-static loading.

\section{Discussion}
\label{sec:discussion}
A central message of this work is that there are two distinct ways to reduce interference with the nominal controller. The first is inherent to the CBF-QP formulation: for a given constraint, the QP leaves a feasible nominal command unchanged and otherwise computes the smallest admissible correction in the chosen input metric. The second is the choice of the constraint itself, whose role can be overlooked when the focus is on minimizing the correction. The constraint determines which commands require intervention in the first place. Here, we use information about the anticipated contact to design that constraint, while retaining a torque interface that can supervise conventional or learned policies without access to their internal structure.

The directional formulation excludes complementary motion and retreat from the kinetic-energy budget for the selected contact. As shown by \eqref{eq:set_inclusion}, this admits states that a total kinetic-energy bound would exclude. Proposition~\ref{prop:velocity} further shows how this difference can translate into higher admissible speeds at the same energy budget. The reaching experiment illustrates the practical consequence: the total bound leaves much of the directional budget unused and produces a substantially slower transient. 

The other motion experiments show why the choice of constraint depends on the contact context. In the $\boldsymbol{d}_y$ reaching trial, little of the nominal motion was directed toward the anticipated contact, and the response remained close to nominal. During circular motion, the filter constrained approach toward $+x$, while retreat and motion along the other axis did not consume the directional budget. This distinction matters in a handover, where the robot may need to limit its approach toward a hand while retaining freedom to make lateral corrections or withdraw. A total bound remains useful when the contact direction is unknown or when the application calls for a global limit.

Two questions remain for future work. First, the experiments use prescribed contact directions, whereas in a shared workspace these directions can change and several potential contacts may need to be considered at once. Future work should examine how uncertainty in online estimates of contact directions and the simultaneous enforcement of several bounds affect filter feasibility and motion freedom. Second, as the circular trial illustrates, a correction that is minimal in torque can still change the end-effector path, including motion along axes other than the protected one. This motivates investigating whether an objective that accounts for the trajectory can better preserve the nominal path while enforcing the same directional kinetic-energy bound.
\section{Conclusion}
\label{sec:conclusion}

We presented a directional kinetic-energy CBF that constrains approach toward a specified potential contact using a dynamically consistent task and null-space decomposition. The resulting torque-affine condition can supervise nominal commands through a standard QP safety filter. Compared with total kinetic-energy limitation, the directional bound admits larger velocities by leaving complementary motion and retreat outside its budget. Four experiments demonstrate the resulting speed advantage, direction-dependent preservation of nominal behavior, and lower transient contact-force peaks. These results support contact-specific kinetic-energy limitation as a means of reducing unnecessary restrictions while retaining a modular interface for conventional and learned controllers.



\bibliographystyle{IEEEtran}
\bibliography{root}
\end{document}